# SDoH-GPT: Using Large Language Models to Extract Social Determinants of Health (SDoH)

Bernardo Consoli[1,2], Xizhi Wu[3], Song Wang[1], Xinyu Zhao[4], Yanshan Wang[3], Justin Rousseau[6], Tom Hartvigsen[7], Li Shen[5], Huanmei Xu[8], Yifan Peng[9], Qi Long[5], Tianlong Chen[4], Ying Ding[1]

[1]University of Texas at Austin, [2]Pontifical Catholic University of Rio Grande do Sul, [3]University of Pittsburgh, [4]University of North Carolina at Chapel Hill, [5]University of Pennsylvania, [6]University of Texas Southwestern, [7]University of Virginia, [8]Temple University, [9]Weill Cornell Medicine

**Abstract**

Extracting social determinants of health (SDoH) from unstructured medical notes depends heavily on labor-intensive annotations, which are typically task-specific, hampering reusability and limiting sharing. In this study we introduced SDoH-GPT, a simple and effective few-shot Large Language Model (LLM) method leveraging contrastive examples and concise instructions to extract SDoH without relying on extensive medical annotations or costly human intervention. It achieved tenfold and twentyfold reductions in time and cost respectively, and superior consistency with human annotators measured by Cohen's kappa of up to 0.92. The innovative combination of SDoH-GPT and XGBoost leverages the strengths of both, ensuring high accuracy and computational efficiency while consistently maintaining 0.90+ AUROC scores. Testing across three distinct datasets has confirmed its robustness and accuracy. This study highlights the potential of leveraging LLMs to revolutionize medical note classification, demonstrating their capability to achieve highly accurate classifications with significantly reduced time and cost.

## Introduction

Social determinants of health (SDoH) [1-9], contributing to an astonishing 80-90% of health outcomes [10-12]. The coexistence of multiple SDoH factors within individuals can significantly exacerbate health risks [13-16]. Critical SDoHs are locked in unstructured clinical narratives [18 -21]. Methods for SDoH extraction using natural language processing (NLP) encompass rule-based, tool-based, and supervised/unsupervised learning approaches, relying on annotated data and lexicons constructed manually or semi-automatically [22]. This manual procedure depends extensively on guidelines that steer the annotation process [21, 23], typically task-specific, resulting in poor reusability. Large Language Models (LLMs) [24], including pre-trained domain specific LLMs [25], have demonstrated promising potential across various healthcare applications [26-30]. A primary obstacle in classifying SDoH from medical notes is the challenge of obtaining high-quality annotations to train machine learning models. LLMs' potential for data labeling has been explored across various NLP tasks, showing great potential to lower cost and yield results comparable to human annotations [31-35]. In this paper, we introduced a simple and effective few-shot learning LLM method that leverages contrastive examples and concise instructions to extract SDoH without relying on extensive medical annotation guidelines or costly human intervention. Furthermore, we trained XGBoost classifiers using LLM annotated SDoH data to achieve optimal performance with affordable computational resources. XGBoost was chosen over transformer-based Language Models (LM) like BERT because XGBoost is computationally cheaper to run and can achieve great performance once the training dataset growing up to 500 and above. Most language models must be fine-tuned on large relevant texts to achieve superior results in specialized tasks. Given the heterogeneous nature of interinstitutional medical notes, as well as the challenges of sharing medical data across different institutions, XGBoost is far more practical and computationally efficient.

In this study, we developed SDoH-GPT utilizing GPT-3.5 with few-shot learning to extract SDoH from medical notes and tested SDoH-GPT on three datasets: MIMIC-SBDH [37]; the Suicide Notes Dataset from the National Violent Death Reporting System (NVDRS) [38]; and the Sleep Notes dataset from a private medical center [39] (See Method; Extended Data 1-3). SDoH-GPT has achieved comparable accuracy to human annotations with a tenfold reduced time and a twentyfold decrease in cost, for 2048 sample annotations (Fig. 1d). If more annotations had been performed, using SDoH-GPT would have become even cheaper and less time-consuming. When compared to human annotation, SDoH-GPT can become a thousandfold cheaper and a hundredfold faster while maintaining comparable accuracy (Extended Data 4). With different few-shot learning strategies, SDoH-GPT has reached superior consistency with human annotators measured by Cohen's kappa: 0.72 to 0.92 in MIMIC-SDBH, 0.71 to 0.88 in Suicide Notes, and 0.70 to 0.91 in Sleep Notes (Fig.1e). The synergistic integration of SDoH-GPT and XGBoost harnesses the strengths of both sides to ensure high accuracy and computational efficiency.

## Results

### SDoH-GPT: An Effective SDoH Classifier

Training an XGBoost [40] classifier using SDoH-GPT annotations can yield enhanced accuracy, consuming less computational resources (Fig. 1a). The scaling up LLMs can perform agnostic tasks with few-shot or zero-shot learning, surpassing state-of-the-art fine-tuning approaches [41,42]. We selected three SDoH categories from MIMIC-SDBH with the highest lexical diversity: Community (i.e., the presence of active social support), Economics (i.e., current employment status), and Tobacco Use (i.e., current or past tobacco consumption) [37]. Fig.1b shows the average performance of XGBoost classifiers for three MIMIC SDoH categories trained on 16 to 2048 annotated examples either by human annotators or SDoH-GPT with zero-shot learning (called 0-shot SDoH-GPT), indicating that with mere 256 examples, XGBoost classifier trained on SDoH-GPT annotations (called XGBoost-SDoH-GPT) can reach ~0.90 AUROC. The discrepancy between XGBoost-SDoH-GPT and XGBoost classifier trained with human annotations (called XGBoost-Human) with more than 256 annotations was marginal, within a range of 0.014 to 0.022 AUROC (Fig.1b). In scenarios with high lexical diversity, such as Economics, XGBoost-Human trained on 256 annotations achieved 0.95 AUROC, while XGBoost-SDoH-GPT trained on 256 annotations by SDoH-GPT two-shot learning (called 2-Shot SDoH-GPT), consistently maintained above 0.90 AUROC scores (Extended Data 5). For Community, XGBoost-Human trained on 512 annotations surpassed 0.95 AUROC, in contrast to XGBoost-SDoH-GPT on 1024 2-Shot SDoH-GPT annotations to maintain the same AUROC score. However, XGBoost-SDoH-GPT with a mere 256 0-Shot SDoH-GPT annotations achieved 0.90 AUROC. The AUROC difference was within a range of 0.014 to 0.035 when comparing XGBoost-Human with XGBoost-SDoH-GPT trained on 2048 annotations. In Tobacco Use, both XGBoost-Human and XGBoost-SDoH-GPT, trained with 128 annotations, reached above 0.90 AUROC. These results demonstrate the comparable performance of XGBoost-Human and XGBoost-SDoH-GPT (Extended Data 5). Fig.1d illustrates that the average performance of XGBoost-Human and XGBoost-SDoH-GPT, trained on 2048 annotations, shows a negligible difference of 0.0194 in AUROC. However, SDoH-GPT is nearly 19 times more cost-effective and 13 times faster when annotating 2048 samples (see Method; Extended Data 5).

Moreover, annotated examples by human annotators and SDoH-GPT shared extremely high agreement measured by Cohen's kappa [43]. Cohen's kappa score as low as 0.41 is deemed acceptable for health-related studies [43]. In our experiments, Cohen's kappa scores demonstrated significant agreement between human annotators and 2-shot SDoH-GPT: 0.87 for Community, 0.82 for Economics, and 0.92 for Tobacco Use, 0.88 for Suicide Notes, across 1024 annotations, and 0.91 for Sleep Notes, across 236 annotations (Fig.1e). These results indicate strong and near-perfect agreement between human annotators and SDoH-GPT. For comparison, Cohen's kappa between two human annotators conducting similar SDoH annotations on clinic notes from Brigham and Women's Hospital/Dana-Farber Cancer Institute was 0.76 for Employment status (akin to MIMIC Economics) and Social Support (akin to MIMIC Community) [45]. Setting the threshold at a 0.8 Cohen's kappa, all SDoH categories, from MIMIC, Suicide Notes, and Sleep Notes, surpassed this benchmark, underscoring a high level of consistency between human annotators and SDoH-GPT (Fig.1c).

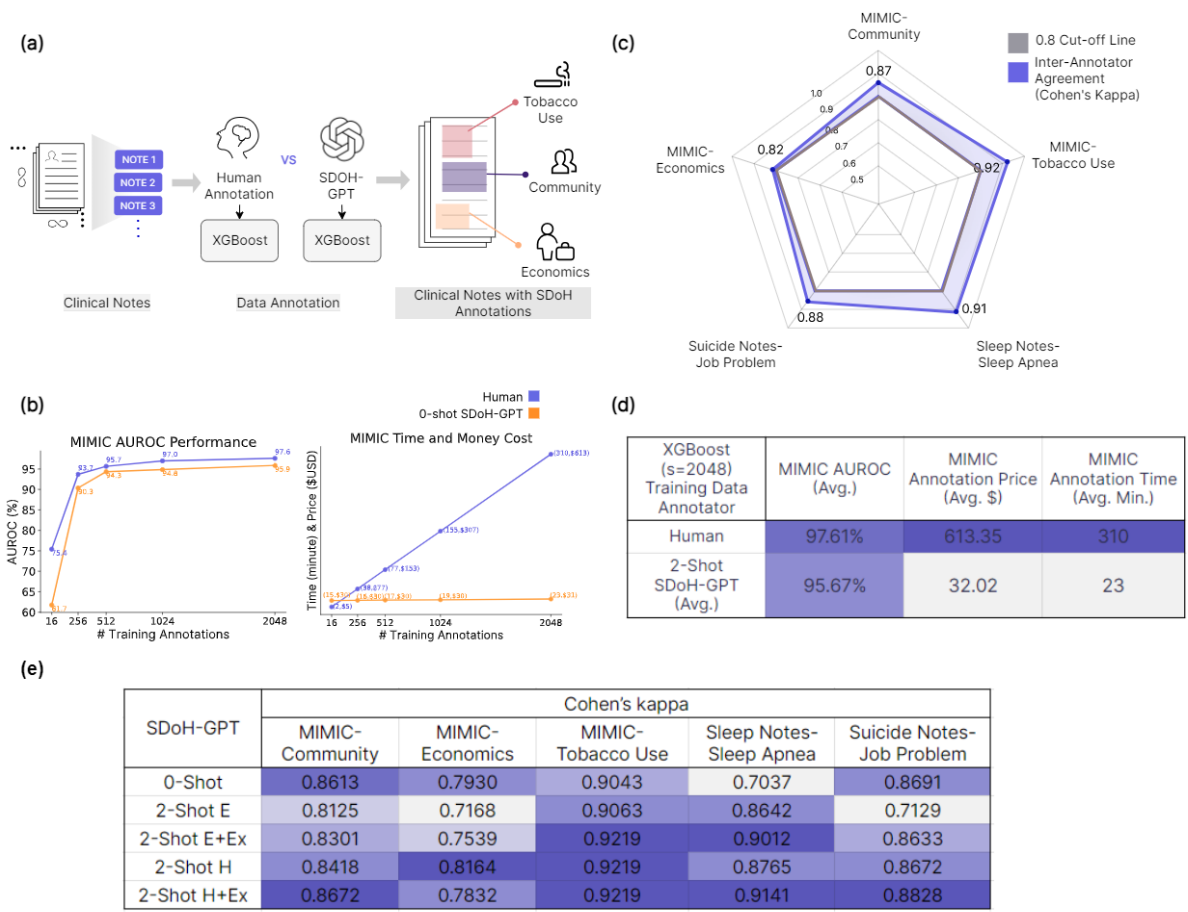

**Figure 1.** Overview of SDoH-GPT for SDoH annotation. (a) Flowchart to show the combination of SDoH-GPT and XGBoost to classify SDoH from clinical notes. (b) Comparison of the average performance, time, and cost of XGBoost classifiers trained on different numbers of human annotations or 0-Shot SDoH-GPT annotations for the three MIMIC SDoH categories (number ranges from 16 to 2048 examples). (c) Spider graph of Cohen's Kappa values for the consistency between human annotations and SDoH-GPT

annotations with the best performed consistency for each SDoH category for MIMIC, Suicide Notes, and Sleep Notes. (d) Table of average performance of XGBoost classifiers trained by 2048 SDoH examples annotated by human or 2-Shot SDoH-GPT, and its corresponding average cost in time and dollars for three MIMIC SDoH categories. (e) Breakdowns of Cohen's Kappa between human, 0-ShotSDoH-GPT, and each 2-Shot SDoH-GPT for 1024 annotated examples from MIMIC-SDBH and Suicide Notes, and 236 annotated examples from Sleep Notes (Extended Data 6).

**SDoH-GPT: A simple few-shot learner**

Fig.2b illustrates the structured template of a prompt, comprising three sections: Instructions, Examples, and Query. The Instructions section includes three components: Roleplaying, which establishes a specific role for GPT-3.5; General Task Instruction, succinctly outlining a goal for GPT-3.5; and SDoH Specific Instruction, providing specific instructions for each SDoH category. This section does not require expertise from domain specialists or intensive details from annotation guidelines. The Examples section includes two examples. The Query section encompasses a medical note (i.e., the social history) and a question regarding the appropriate SDoH category for classification. The combination of examples is thoroughly explained in Fig.2c. First, we deployed 0-Shot SDoH-GPT on 100 randomly chosen discharge summaries from MIMIC-SDBH for each SDoH category. Subsequently, we compared SDoH-GPT annotations with human annotations to identify True Positives, True Negatives, False Positives, and False Negatives. This led to the formulation of four two-example strategies for the Examples section: combining one positive example from True Positives and one negative example from True Negatives to create 2-Shot E, with added explanations forming 2-Shot E+Ex; and selecting one positive example from False Negatives and one negative example from False Positives to develop 2-Shot H, with explanations yielding 2-Shot H+Ex. Our experiment with different numbers of examples in prompts, ranging from zero to eight, revealed that prompts with zero and two examples exhibit comparable accuracy to those with eight examples (Fig.3d). Our prompt template is simpler than those from [36] which provide an extensive human-generated annotation guideline (Extended Data 7-8). [46] used GPT with zero-shot learning to annotate SDoH on 1,000 medical notes from a University Hospital. Their GPT template is simple, but their F1 scores for Employment status (0.613), Tobacco use (0.652), and Living status (0.608) are much lower than ours (Fig.3d).

Fig.2a shows the workflow of SDoH-GPT. In the Data Extraction section, we prepared the Training Set with 1024 positive and 1024 negative examples; and the Testing Set with 512 positive and 512 negative examples for each MIMIC SDoH category. Then we ran 0-Shot SDoH-GPT on 100 randomly selected examples from the Training Set to select two examples for 2-Shot SDoH-GPT. We utilized Regular Expressions (RegEX) to extract the Social History subsection from discharge summaries of MIMIC-III, excluding neonates and those with missing values. This process yielded 37,558 social history notes without human annotations (called Unannotated Social Histories). Then we ran 0-Shot SDoH-GPT and 2-Shot SDoH-GPT on the Unannotated Social Histories to produce 1024 positive and 1024 negative examples for each SDoH category. We have six training sets with 1024 positive and 1024 negative examples for each MIMIC SDoH category: Human-Annotated Training Set (same as the Training Set), 0-Shot SDoH-GPT Training Set, 2-Shot E SDoH-GPT Training Set, 2-Shot E+Ex SDoH-GPT Training Set, 2-Shot H SDoH-GPT Training Set, and 2-Shot H+Ex SDoH-GPT Training Set. We trained six XGBoost classifiers, one for each Training Set, and evaluated their performance using AUROC based on the Testing Set.

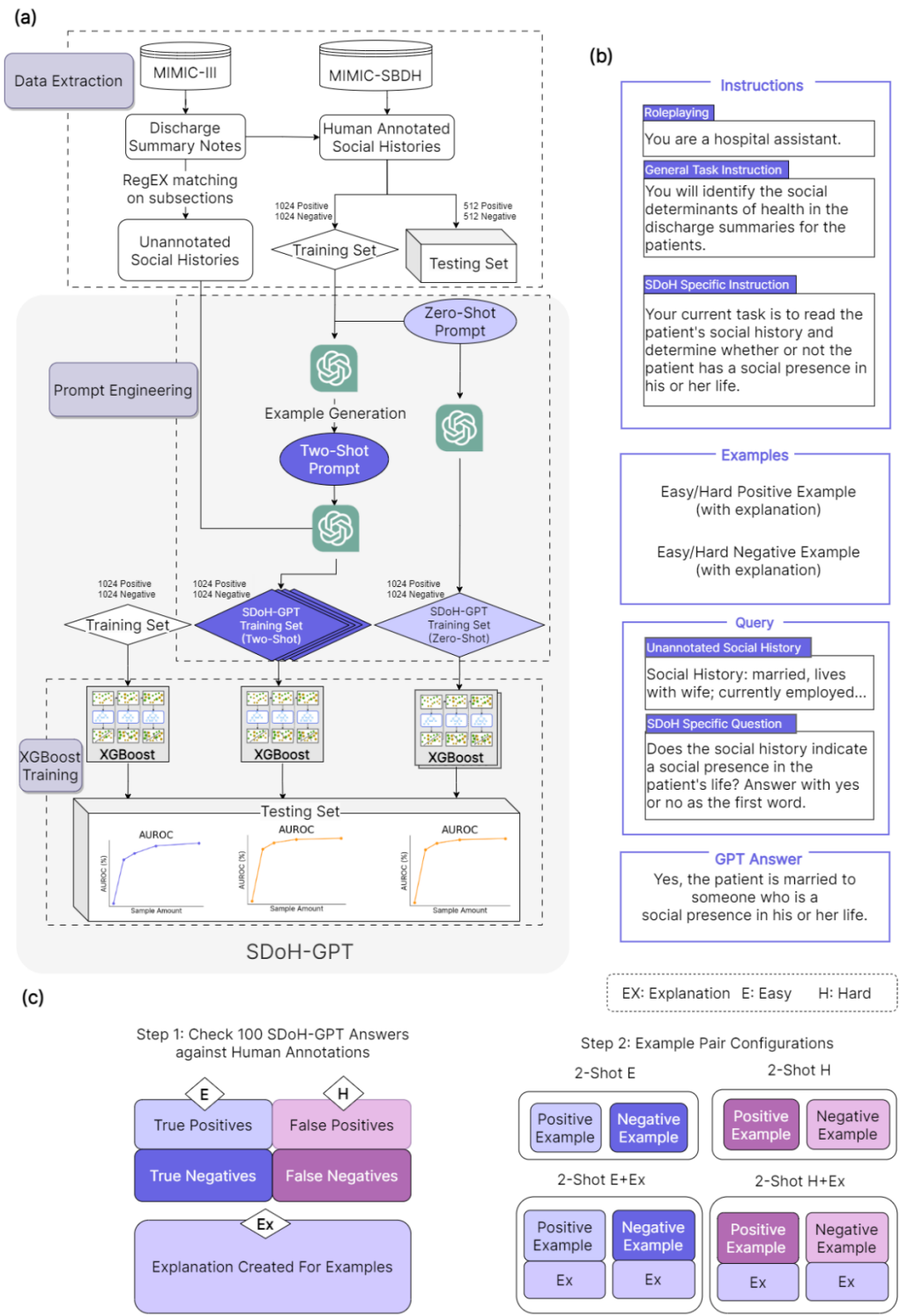

**Figure 2.** The workflow of SDoH-GPT. (a) Three sections on how to build SDoH-GPT for three MIMIC SDoH categories: Data Extraction, Prompt Engineering, and XGBoost Training. (b) Three sections of SDoH-GPT

prompt template: Instructions, Examples, and Query. (c) Two steps on how to generate examples for 2-Shot SDoH-GPT.

**SDoH-GPT: Classifying SDoH for MIMIC Discharge Summaries**

We evaluated SDoH-GPT on MIMIC-III discharge summaries to classify the top three lexically diverse SDoH categories: Community, Economics, and Tobacco Use. It is crucial to consolidate the various SDoH category values into binary classifications of 'Yes' or 'No' (Fig.3a). For example, MIMIC Tobacco Use was simplified from five to two binary values: 'Yes' encompassing both Past and Present usage, and 'No' indicating either No Mention or Never Used, while Unsure was removed. To counteract data imbalance, a random sampling method was employed to balance the number of positive and negative examples. Next, the balanced annotation set was divided into a Training Set with 1024 positive and 1024 negative examples, and a Testing Set with 512 positive and 512 negative examples (Fig.3a). Fig.3b shows the comparative performance of XGBoost classifiers trained on the annotations of best performing SDoH-GPT for each category (2-Shot H+Ex SDoH-GPT for Community; 0-Shot SDoH-GPT for Economics; 2-Shot E SDoH-GPT for Tobacco Use) versus those classifiers trained on human annotations, including the associated time and financial costs. Given the high expense and complexity inherent in human annotation, related studies typically have a limited number of SDoH annotations: 1,000 [46], 1,576 [47] and 500 [48]. Assuming only 512 human annotations are available for Community, SDoH-GPT can effortlessly generate 2048 additional annotations at an additional cost of $1.89 and an extra 8.5 minutes, achieving a 0.01 higher AUROC score compared to XGBoost trained on 512 human annotations. Similarly, for Economics, SDoH-GPT could produce 2048 additional annotations at an additional $0.93 and 8.5 extra minutes, attaining the same AUROC score. Likewise, for Tobacco Use, SDoH-GPT could generate 2048 additional annotations with an increase of $1.87 and an additional 8.5 minutes, resulting in a 0.018 higher AUROC score (Fig.3b). Notably, SDoH-GPT requires only 100 human-annotated examples in total to select two examples for 2-Shot SDoH-GPT, demonstrating significant cost-effectiveness and efficiency. Ablation studies revealed minimal variance in AUROC scores between 0-Shot SDoH-GPT and various 2-Shot SDoH-GPT trained on 2048 annotations, suggesting that additional shots do not necessarily enhance performance (Fig.3c). Employing 2-Shot SDoH-GPT directly for annotating the Testing Set (512 positive and 512 negative examples) without XGBoost yielded an F1 score nearly equivalent to that achieved by using XGBoost trained on 2048 human annotations (Fig.3d). This result indicates that SDoH-GPT can significantly reduce manual annotation efforts, needing twenty times fewer human annotations to maintain comparable F1 scores (Fig.3d). 0-Shot SDoH-GPT on the Testing data outperformed XGBoost classifier trained on 2048 human annotations in terms of F1 scores in all categories except Economics, without necessitating any human annotations (Fig.3d).

In this study, we attained a notably higher F1 score using 2-Shot SDoH-GPT: 0.905 in Economics, which is 0.102 higher than those from GPT-4 in [36]; 0.963 in Tobacco Use, surpassing 0.138 than those in [36]; and 0.926 in Community, a significant improvement of 0.336 over Living Status in [36]. Several factors potentially contribute to the differences: 1) Variance in GPT prompt structure: [36]'s query in GPT prompts was to annotate discharge summaries in the BRAT standoff format, while ours were pure SDoH categorization which are triggers in BRAT; 2) Dataset differences: [36] contains MIMIC III and an additional dataset from the University of Washington; while ours are MIMIC III and two other datasets 3) Instruction guideline: [36]'s prompt included a lengthy instruction, exceeding 1,000 characters; while our instruction employed a more concise instruction, limited to a few sentences; and 4) Two-shot learning: [36]'s prompt did not have two examples to facilitate two-shot learning. Moreover, [45] involved a manual annotation of 200 medical notes from MIMIC-III and finetuned Flan-T5 containing 18 million parameters, using a parameter-efficient fine-tuning method called LoRA. It reported 0.44 F1 in Community and 0.55 in Economics for MIMIC III. However, our F1 scores in these two categories notably doubled, an enhancement attributable to our SDoH-GPT uses GPT 3.5 which has a higher number of parameters than Flan-T5. Assuming only 256 human annotations are available, SDoH-GPT can effortlessly generate more annotations and XGBoost-SDoH-GPT can increase AUROC scores with up to 0.04 improvement (Fig.3e).

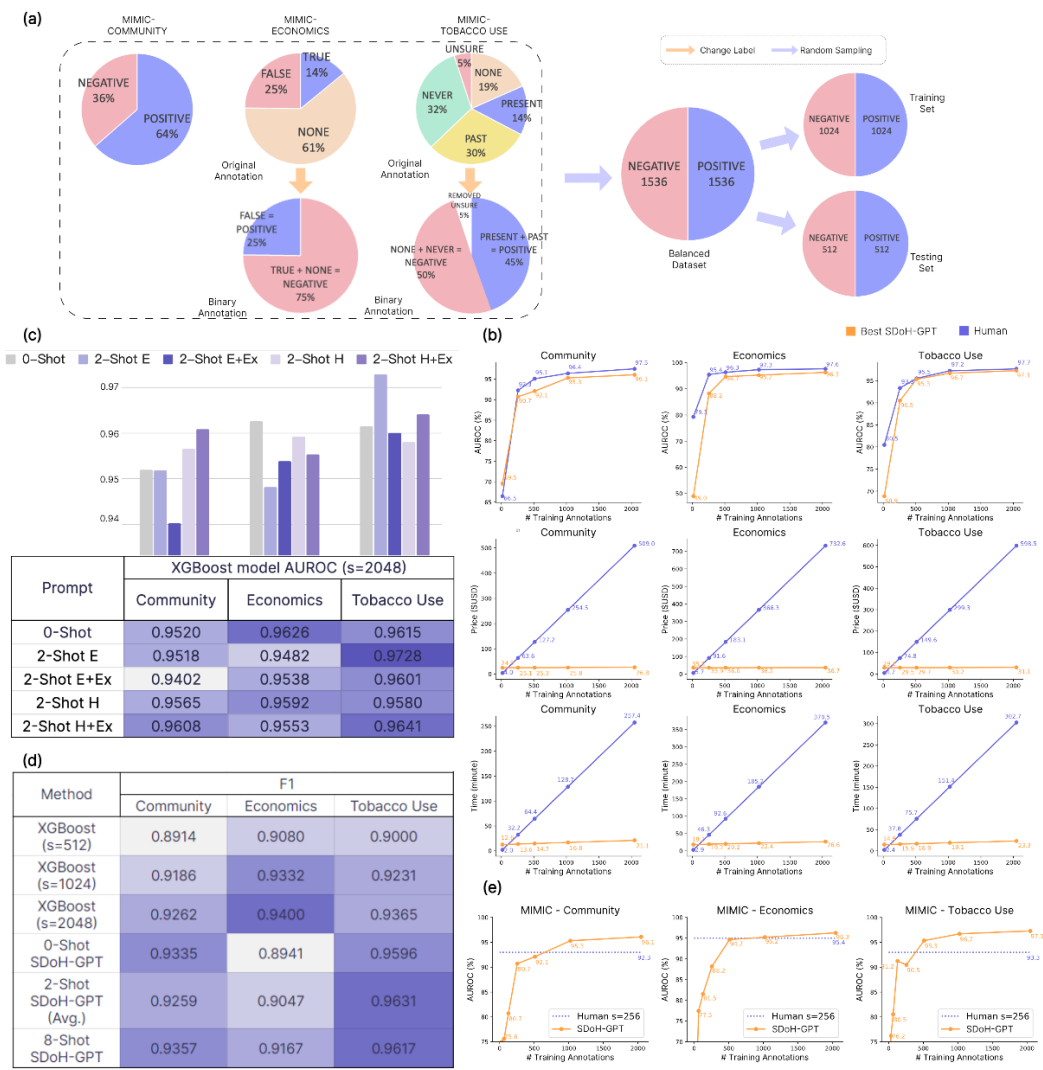

**Figure 3.** SDoH-GPT for MIMIC SDoH Classification. (a) Merging different values of each MIMIC SDoH category into balanced binary values to create Training Set with 1024 positive examples and 1024 negative examples and Testing Set with 512 positive examples and 512 negative examples. (b) The comparison results of AUROC, time and money cost for XGBoost classifiers trained on human annotated examples versus those trained on annotated examples from 2-Shot H+Ex SDoH-GPT for Community, the 0-Shot SDoH-GPT for Economics, and 2-Shot E SDoH-GPT for Tobacco Use (the best performing SDoH-GPT for their respective tasks). (c) Ablation studies to show the AUROC performance of XGBoost classifier trained on 2048 examples generated by 0-Shot SDoH-GPT and different 2-shot SDoH-GPT. (d) Comparison of F1 measures on directly using Zero-Shot and Few-Shot SDoH-GPT to classify three MIMIC SDoH categories of the Testing Set (1024 samples) without training XGBoost classifiers, against XGBoost classifiers directly trained on different numbers of human annotated examples. (e) Assuming only 256 human annotations are available (the dotted line), the number of 2-Shot H+Ex SDoH-GPT annotated examples are needed to reach the same level of accuracy for Community; same for Economics with 0-Shot SDoH-GPT and Tobacco Use with 2-Shot E SDoH-GPT. Notably for AnnotateGTP, this number of annotations require less than 10 minutes and cost less than 2 dollars' worth of computation.

**Validating SDoH-GPT using Suicide Notes and Sleep Notes**

SDoH-GPT was validated using two distinct datasets: Suicide Notes and Sleep Notes. This validation followed the methodology delineated in Fig.2a, which guided the creation of both Training and Testing Sets (Fig.4b). For Sleep Notes, each note was segmented into multiple sentences and each sentence was annotated using all five SDoH-GPT prompts to detect sleep apnea. For Suicide Notes, each note comprised two specific sections, namely the coroner or medical examiner's report and the law enforcement report, which were considered to comprise one medical note. All 0-Shot SDoH-GPT and various 2-Shot SDoH-GPT were employed to ascertain the association between job problems and the victim's suicide ideation (Fig.4a). Fig.4c presents the results in AUROC, time and cost. For Suicide Notes, XGBoost-SDoH-GPT trained on annotations from 2-Shot H SDoH-GPT performs comparably to XGBoost-Human with significantly reduced time and computational cost. If only 256 human annotations are available, XGBoost-SDoH-GPT exhibited a substantial increase in AUROC score by 0.036 (Fig.4d). Moreover, with mere 128 annotations, both XGBoost-Human and XGBoost-SDoH-GPT with 0-Shot, 2-Shot E+Ex and 2-Shot H+Ex attained 0.90 AUROC score. However, human annotations incurred greater financial and time costs (Extended Data 5). In scenarios where only 512 human annotations were feasible, XGBoost trained on 2048 2-shot H SDoH-GPT annotations exhibited a remarkable enhancement in AUROC scores, escalating from 0.946 to 0.973. This is particularly noteworthy as the time and cost associated with SDoH-GPT remain constant, whereas those for human demonstrate significant escalations. For Sleep Notes, where only 236 human annotations were available for sleep apnea (74 as Training Set and 162 as Testing Set), the peak AUROC achieved by XGBoost-Human is 0.922. By spending an extra $2.68 dollars and 8.5 minutes, XGBoost-SDoH-GPT, trained 2048 annotations generated by 2-Shot H SDoH-GPT, can improve AUROC from 0.922 to 0.964 (Fig.4c, Fig.4d). This demonstrates the substantial utility of SDoH-GPT in areas where human annotations are scarce and expensive to obtain. SDoH-GPT can markedly enhance performance with minimal additional effort.

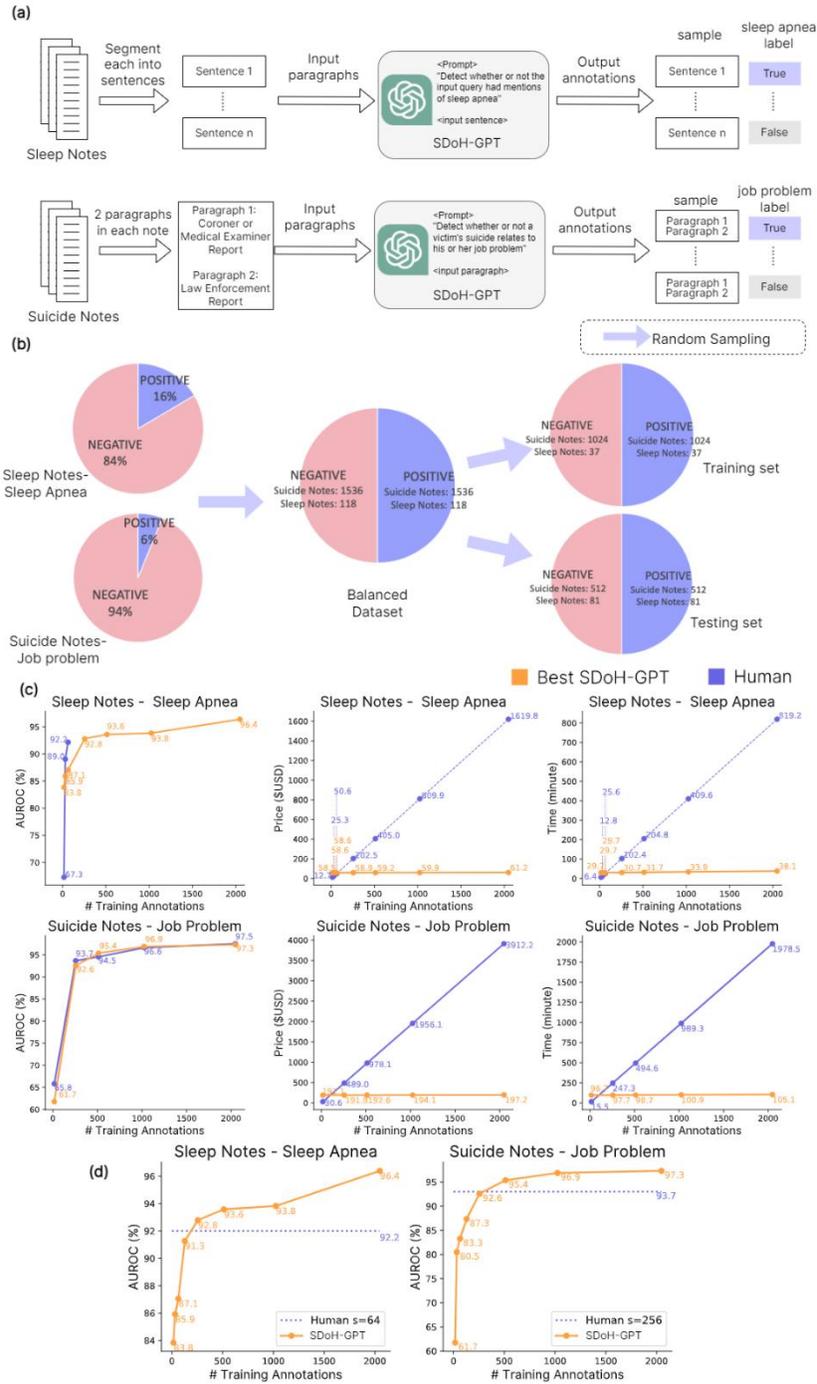

**Figure 4.** SDoH-GPT Evaluation on Sleep Notes and Suicide notes. (a) Data pre-processing. (b) Procedure to generate the Training and Testing Sets. We took a large set for Testing Set for Sleep Notes to ensure stable performance scores to avoid huge variances due to the small samples of Testing Set. (c) Comparable performance of XGBoost-Human versus XGBoost-SDoH-GPT trained on different numbers of annotations generated by either human annotators or 2-Shot H SDoH-GPT (the best performing SDoH-GPT for both validation tasks). (d). Performance improvement by XGBoost-SDoH-GPT for use cases where human annotations are limited.

**Understanding SDoH-GPT Classification Errors**

We employed SDoH-GPT to classify the Testing Set in MIMIC-SDBH (Fig.2a) to conduct a thorough error analysis. Fig.5a and 5b show four categories of errors: Human error (i.e., errors in human annotations); SDoH-GPT error (i.e., errors in SDoH-GPT annotations); and Extraction error (i.e., incorrect extractions of social histories from discharge summaries using Regular Expression algorithms), and Ambiguity (i.e., hard to decide). These are several instances of annotation ambiguity (Fig.5c): 1) Contextual Misinterpretations: human annotators classified Case1 as community absent, whereas SDoH-GPT identified it as community present, misinterpreting the context of "lost family". 2) Temporal Conditions: human annotators mistakenly treated the subject in Case2 as currently employed, neglecting past tense and context of having five great-grandchildren. Conversely, SDoH-GPT categorized this as unemployed or retired. 3) Evolving Status: SDoH-GPT's annotation in Case3 was limited to the initial part of the sentence, leading to the annotation of community present. However, considering a potential 5-day rehabilitation stay, his situation could shift to community present within five days, indicating a temporary condition. 4) Implicit Statements: Case4 suggests daily access to community services for the patient, yet this does not explicitly imply anything about community presence, while SDoH-GPT marked it as community present. Daily access to healthcare services cannot directly infer community presence [52]. 5) Incomplete Information: It is unclear whether the patient in Case5 lives with his children or not. The patient's searching for housing could imply a lack of permanent residence, rendering his community presence ambiguous. Nevertheless, SDoH-GPT classified this as community present, which can be questionable. An extreme example illustrates the challenges in interpretation of "A patient has never smoked and drinks socially alcohol" as community present by SDoH-GPT, presumably due to the social implication of alcohol consumption. Fig.5d demonstrates the inconsistencies in annotations between human annotators and SDoH-GPT for roughly the same social histories across two visits of the same patient.

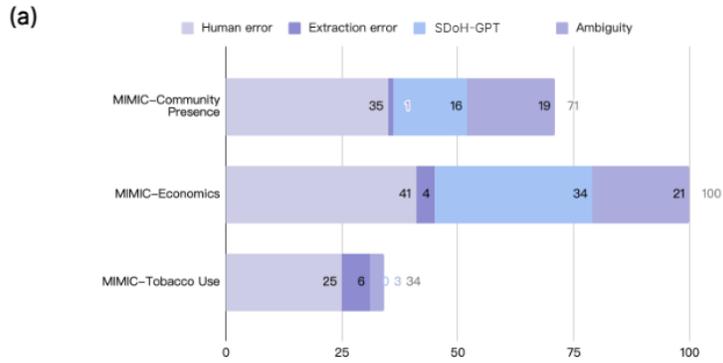

**Figure 5.** SDoH-GPT errors. (a) Statistics of SDoH-GPT error types for each MIMIC SDoH category. (b) Examples for each SDoH-GPT error type for MIMIC SDoH categories. (c) Ambiguity cases and their corresponding human annotations and SDoH-GPT annotations. (d) Inconsistency for human annotations and SDoH-GPT annotations for a patient's two visits. All example notes are recreated.

## Discussion

SDoH data in medical notes are concise, lacking detailed context. This brevity often results in implicit statements, which introduces ambiguity into SDoH annotations. There are several pivotal issues concerning ambiguity: 1) Overgeneralization: LLMs are predominantly trained on patients in urban areas, leading to inaccurate extrapolations for rural patients [53]. 2) Individualization: The living conditions of a patient in an elder care facility can lead to diverse SDoH annotations [52]. 3) Incompleteness: The absence of comprehensive information further compounds the ambiguity in determining SDoH status [54]. 4) Misclassification: LLMs often classify mentions of "community support" as references to social activities, overlooking contexts where it might pertain to community healthcare services [55]. Ambiguity in SDoH stems from its inherently complex and multifaceted nature, requiring a nuanced understanding and context-specific analysis to ensure precise and meaningful annotation [47, 50, 56]. Effectively addressing these issues requires a comprehensive understanding of the medical domain, as well as the broader societal context pertinent to healthcare. Our proposed approach, which employs XGBoost trained with SDoH-GPT annotations, presents an efficient and straightforward solution. This approach attains accuracy comparable to human-level performance for SDoH annotation while requiring fewer computational resources. Nevertheless, our paper has certain limitations. First, our SDoH categorization is binary (Yes or No), which may not adequately capture clinical complexity. Second, our current experiment was limited to the categories of Community, Economics, and Tobacco Use, excluding other SDoH categories. Last, our approach to SDoH annotation is confined to the categorical level and does not extend to sentence-level annotation of triggers and spans.

## Data Availability

MIMIC-III (https://physionet.org/content/mimiciii/1.4/) and NVDRS (https://www.cdc.gov/violenceprevention/datasources/nvdrs/dataaccess.html) are third-party datasets available for credentialed use. MIMIC-SBDH is a publicly available third-party dataset (https://github.com/hibaahsan/MIMIC-SBDH). The Sleep Notes dataset is protected by HIPAA law. It is restricted to research and healthcare use. Access can be granted by the University of Pittsburgh's Office of Sponsored Programs (osp@pitt.edu), which has a 2–6-month response timeframe.

## Code Availability

Our codebase is available upon request (corresponding author).

## Reference

1. Virnig, B.A., Baxter, N.N., Habermann, E.B., Feldman, R.D. & Bradley, C.J. A matter of race: early-versus late-stage cancer diagnosis. Health Affairs 28, 160–168 (2009).
2. Özdemir, B.C. & Dotto, G.P. Racial differences in cancer susceptibility and survival: more than the color of the skin?. Trends in Cancer 3, 181–197 (2017).
3. American Cancer Society. Cancer Facts & Figures 2020 (American Cancer Society, 2020).
4. Burgard, S.A., Ailshire, J.A., & Kalousova, L. The great recession and health: people, populations, and disparities. The Annals of the American Academy of Political and Social Science 650, 194-213 (2013).

5. Draper C.E., Grobler L., Micklesfield L.K. & Norris S.A. Impact of social norms and social support on diet, physical activity and sedentary behaviour of adolescents: a scoping review. Child: Care, health and development 41, 654-667 (2015).
6. Szreter, S., & Woolcock, M. Health by association? Social capital, social theory, and the political economy of public health. International Journal of Epidemiology 33, 650-667 (2004).
7. Williams, D.R., & Mohammed, S.A. Discrimination and racial disparities in health: Evidence and needed research. Journal of Behavioral Medicine 32, 20-47 (2009).
8. Chen M., Tan X. & Padman R. Social determinants of health in electronic health records and their impact on analysis and risk prediction: a systematic review. Journal of the American Medical Informatics Association 27, 1764-1773 (2020).
9. Hatef E. et al. Assessing the availability of data on social and behavioral determinants in structured and unstructured electronic health records: a retrospective analysis of a multilevel health care system. JMIR medical informatics 7, e13802 (2019).
10. Braveman P. & Gottlieb L. The social determinants of health: it's time to consider the causes of the causes. Public Health Reports 129, 19-31 (2014).
11. Magnan S. Social determinants of health 101 for health care: five plus five. NAM Perspectives, https://doi.org/10.31478/201710c (2017).
12. Hood, C.M., Gennuso, K.P., Swain, G.R., & Gatlin, B.B. County health rankings: Relationships between determinant factors and health outcomes. American journal of preventive medicine 50, 129-135 (2016).
13. Bejan, C.A. et al. Mining 100 million notes to find homelessness and adverse childhood experiences: 2 case studies of rare and severe social determinants of health in electronic health records. Journal of the American Medical Informatics Association 25, 61–71 (2017).
14. Schroff P., Gamboa C.M., Durant R.W., Oikeh A., Richman J.S., Safford M.M. Vulnerabilities to health disparities and statin Use in the REGARDS (Re asons for Geographic and Racial Differences in Stroke) Journal of the American Heart Association 6, e005449 (2017).
15. Reshetnyak E., et al. Impact of multiple social determinants of health on incident stroke. Stroke 51, 2445–2453 (2020).
16. Pinheiro L.C., Reshetnyak E., Sterling M.R., Levitan E.B., Safford M.M. & Goyal P. Multiple vulnerabilities to health disparities and incident heart failure hospitalization in the REGARDS study. Circulation: Cardiovascular Quality and Outcomes 13, e006438 (2020).
17. Pinheiro L.C., Reshetnyak E., Akinyemiju T., Phillips E., Safford M.M. Social determinants of health and cancer mortality in the Reasons for Geographic and Racial Differences in Stroke (REGARDS) cohort study. Cancer 128, 122-130 (2022).
18. Wang, M., Pantell, M.S., Gottlieb, L.M., & Adler-Milstein, J. Documentation and review of social determinants of health data in the EHR: measures and associated insights. Journal of the American Medical Informatics Association 28, 2608-2616 (2021).
19. Spasic I, Nenadic G. Clinical text data in machine learning: systematic review. JMIR Med Informatics 8, e17984 (2020).
20. Shrank, W.H., Rogstad, T.L., & Parekh, N. Waste in the US health care system: Estimated costs and potential for savings. JAMA, 322, 1501-1509 (2019).
21. Wei, Q., Franklin, A., Cohen, T. & Xu, H. Clinical text annotation: What factors are associated with the cost of the time? AMIA Annual Symposium Proceedings 2018, 1552-1560 (2018).

22. Patra B.G. et al. Extracting social determinants of health from electronic health records using natural language processing: a systematic review. Journal of the American Medical Informatics Association 28, 2716-2727 (2021).
23. Fort K., Nazarenko A. & Rosset S. Modeling the complexity of manual annotation tasks: a grid of analysis. International Conference on Computational Linguistics 2012, 895–910 (2012).
24. Achiam, J., et al. GPT-4 Technical Report. Preprint at https://arxiv.org/abs/2303.08774 (2023).
25. Singhal, K., et al. Large language models encode clinical knowledge. Nature 620, 172-180 (2022).
26. Jiang, L.Y. et al. Health system-scale language models are all-purpose prediction engines. Nature 619, 357–362 (2023).
27. Nazario-Johnson, L., Zaki, H.A. & Tung, G.A. Use of large language models to predict neuroimaging. Journal of the American College of Radiology 20, 1004-1009 (2023).
28. Sorin, V., Barash, Y., Konen, E. & Klang, E. Journal of Cancer Research and Clinical Oncology 149, 9505–9508 (2023).
29. Gilson A., et al. How Does ChatGPT Perform on the United States Medical Licensing Examination? The Implications of Large Language Models for Medical Education and Knowledge Assessment. JMIR Medical Education 9, e45312 (2023).
30. Brin, D. et al. Comparing ChatGPT and GPT-4 performance in USMLE soft skill assessments. Scientific Reports 13, 16492 (2023).
31. Ding, B., et al. Is gpt-3 a good data annotator? Preprint at https://arxiv.org/abs/2212.10450 (2022).
32. Guo, X. & Yiqiang C. Generative AI for Synthetic Data Generation: Methods, Challenges and the Future. Preprint at https://arxiv.org/abs/2403.04190 (2024).
33. Agrawal, M., Hegselmann, S., Lang, H., Kim, Y., & Sontag, D. Large language models are few-shot clinical information extractors. Conference on Empirical Methods in Natural Language Processing 2022, 1998-2022 (2022).
34. Sushil, M., Kennedy, V. E., Mandair, D., Miao, B. Y., Zack, T. & Butte, A. J. CORAL: Expert-Curated medical Oncology Reports to Advance Language Model Inference. Preprint at https://arxiv.org/abs/2308.03853 (2023).
35. Goel, A. et al. LLMs accelerate annotation for medical information extraction. Machine Learning for Health 225, 82-100 (2023).
36. Ramachandran, G.K. Prompt-based Extraction of Social Determinants of Health Using Few-shot Learning. Proceedings of the 5th Clinical Natural Language Processing Workshop 5, 385–393 (2023).
37. Ahsan H, Ohnuki E, Mitra A & Yu H. MIMIC-SBDH: A Dataset for Social and Behavioral Determinants of Health. PMLR 149, 391-413 (2021)
38. Liu G.S., et al. Surveillance for Violent Deaths - National Violent Death Reporting System, 48 States, the District of Columbia, and Puerto Rico, 2020. MMWR Surveillance Summaries 72, 1-38 (2023).
39. Mohammad H.A., et al. Extraction of Sleep Information from Clinical Notes of Alzheimer's Disease Patients Using Natural Language Processing. Preprint at https://arxiv.org/abs/2204.09601 (2022).
40. Chen, T. & Guestrin, C. XGBoost: A scalable tree boosting system. ACM SIGKDD International Conference on Knowledge Discovery and Data Mining 22, 785–794 (2016).
41. Brown, T.B., et al. Language models are few-shot learners. Advances in neural information processing systems 33, 1877-1901 (2020).
42. Nori, H., King, N., McKinney, S.M., Carignan, D. & Horvitz, E. Capabilities of GPT-4 on medical challenge problems. Preprint at https://arxiv.org/abs/2303.13375 (2023).

43. Cohen, J. A coefficient of agreement for nominal scales. Educational and psychological measurement 20, 37-46  (1960).
44. McHugh M.L. Interrater reliability: the kappa statistic. Biochemia medica 22, 276-282 (2012).
45. Guevara, M., et al. Large language models to identify social determinants of health in electronic health records. npj Digital Medicine 7, 6 (2024).
46. Bhate, N.J., Mittal, A., He, Z. & Luo, X. Zero-shot learning with minimum instruction to extract social determinants and family history from clinical notes using GPT model. Preprint at https://arxiv.org/pdf/2309.05475 (2023)
47. Lituiev D.S., Lacar B., Pak S., Abramowitsch P.L., De Marchis E.H., Peterson T.A. Automatic extraction of social determinants of health from medical notes of chronic lower back pain patients. Journal of the American Medical Informatics Association 2023 30, 1438-1447 (2023).
48. Yu Z., et al. A study of social and behavioral determinants of health in lung cancer patients using transformers-based natural language processing models. AMIA Annual Symposium Proceedings 2021, 1225–1233 (2021).
49. Social Security Disability Insurance (SSDI); https://www.ssa.gov/benefits/disability/, (2024)
50. Lybarger, K., Ostendorf, M. & Yetisgen, M. Annotating social determinants of health using active learning, and characterizing determinants using neural event extraction. Journal of Biomedical Informatics 113, 103631 (2021)
51. Andermann A. Taking action on the social determinants of health in clinical practice: a framework for health professionals. CMAJ 188, 474-483 (2016)
52. Boamah S.A., Weldrick R., Lee T.J. & Taylor N. Social isolation among older adults in long-term care: A scoping review. Journal of Aging and Health. 233, 618-632 (2021).
53. Chen X., et al. Differences in rural and urban health information access and use. The Journal of Rural Health 35, 405-417 (2019)
54. Portacolone E., et al. Perceptions of the Role of Living Alone in Providing Services to Patients With Cognitive Impairment. JAMA Network Open 6, e2329913 (2023).
55. Singh R., et al. Community and social context: an important social determinant of cardiovascular disease. Methodist Debakey Cardiovascular Journal 17, 15-27 (2021).
56. Feller DJ, et al. Detecting social and behavioral determinants of health with structured and free-text clinical data. Applied clinical informatics 11, 172–181 (2020).
57. Han S., et al. Classifying social determinants of health from unstructured electronic health records using deep learning-based natural language processing. Journal of Biomedical Informatics 127, 103984 (2022).